\documentclass{article}
\usepackage{graphicx}

\PassOptionsToPackage{square,numbers}{natbib}

     \usepackage[final]{neurips_2022}

\usepackage[utf8]{inputenc} 
\usepackage[T1]{fontenc}    
\usepackage{hyperref}       
\usepackage{url}            
\usepackage{booktabs}       
\usepackage{amsfonts}       
\usepackage{nicefrac}       
\usepackage{microtype}      
\usepackage{xcolor}         

\title{UATTA-ENS: Uncertainty Aware Test Time Augmented Ensemble for PIRC Diabetic Retinopathy Detection}

\author{
  Pratinav Seth \thanks{Manipal Institute of Technology, Manipal Academy of Higher Education, Manipal, India}\\
  Dept. of Data Science and Computer Applications\\
  \texttt{seth.pratinav@gmail.com} \\
   \And
   Adil Khan \footnotemark[1]\protect\phantom{\footnotesize 1}\footnotemark[2]\\
   Dept. of Mechanical Engineering\\
   \texttt{adilk5020@gmail.com} \\
   \And
   Ananya Gupta \footnotemark[1]\protect\phantom{\footnotesize 1}\footnotemark[2]\\
   Dept. of Electronics and Comm. Engineering \\
   \texttt{ananyag1018@gmail.com} 
   \And
     Saurabh Kumar Mishra \footnotemark[1]\protect\phantom{\footnotesize 1}\thanks{Equal contribution.}  \\
  Dept. of Computer Science and Engineering \\
   \texttt{saurabhskm22@gmail.com}
   \And
   Akshat Bhandari \footnotemark[1]\\
   Dept. of Computer Science and Engineering\\
   \texttt{akshatbhandari15@gmail.com} \\
}
\begin{document}
\maketitle
\begin{abstract}
Deep Ensemble Convolutional Neural Networks has become a methodology of choice for analyzing medical images with a diagnostic performance comparable to a physician, including the diagnosis of Diabetic Retinopathy. However, commonly used techniques are deterministic and are therefore unable to provide any estimate of predictive uncertainty. Quantifying model uncertainty is crucial for reducing the risk of misdiagnosis. A reliable architecture should be well-calibrated to avoid over-confident predictions. 
To address this, we propose a UATTA-ENS: Uncertainty-Aware Test-Time Augmented Ensemble Technique for 5 Class PIRC Diabetic Retinopathy Classification to produce reliable and well-calibrated predictions. Implementation is available at  \url{https://github.com/ptnv-s/UATTA-ENS}.
\end{abstract}

\section{Introduction and Previous Works}
In recent years, there has been rapid advancement in the field of Deep Learning; subsequently, enormous headway has been made in applying computer vision techniques to medical imaging \cite{Gondal2017}\cite{8869883}. Different algorithms have been developed and fine-tuned for various diseases \cite{mishra2021use}\cite{Karki2021DiabeticRC}. A critical problem in biomedical image analysis is the occurrence of batch effects \cite{leek2010tackling}, which are differences introduced through technological artifacts among different subsets of data. Sample handling and data collecting techniques make applying computer vision algorithms to data from different pathological labs problematic, which is an essential step in developing ML models.

Diabetic Retinopathy (DR) is a common disease that causes vision loss or, in some cases, even blindness among people with diabetes, affecting 415 million people \cite{Sabanayagam2016TenET}. Automated detection of DR is essential to limit the progression of DR by conducting early diagnoses on diabetic patients, as the normal procedure by ophthalmologists demands many resources.
Therefore, the interest in employing deep neural networks to automatically classify Diabetic Retinopathy has grown over the past few years \cite{Gulshan2016DevelopmentAV}\cite{Gargeya2017AutomatedIO}.

A well-calibrated classifier would place less probability mass on uncertain classes. The issue of uncertainty estimation is especially important in the medical domain in order to trust confident model predictions for screening automation and referring uncertain cases for manual intervention of a medical expert \cite{rahaman2021uncertainty}. Bayesian probability theory offers a sound mathematical framework to design machine learning models with an inherent and explicit notion of uncertainty \cite{Yang2021UncertaintyQA}. Instead of resulting in a single per-class probability, such models can estimate the moments of the output distribution for every class, including mean and variance.

Multiple probabilistic and Bayesian methods such as \cite{Graves2011PracticalVI}\cite{pmlr-v37-blundell15}\cite{HernndezLobato2015ProbabilisticBF}\cite{Blum2015VariationalDA}\cite{Gal2016DropoutAA}\cite{Lee2018DeepNN}\cite{Wu2019DeterministicVI}\cite{Pearce2020UncertaintyIN} and non-Bayesian methods such as \cite{Osband2016RiskVU}\cite{Sarawgi2020WhyHA}\cite{Lakshminarayanan2017SimpleAS} and \cite{Dusenberry2020AnalyzingTR} have been proposed to quantify the uncertainty estimates. The ensemble of networks can further improve the performance of models. There has been some work on incorporating uncertainty where ensemble methods \cite{Kendall_2018_CVPR} learned multiple tasks by using the uncertainty predicted as weights for the losses of each of the models, thus outperforming individual models trained on each task. Non-Bayesian alternatives can offer simpler yet effective means to quantify the uncertainties of DNNs. While the ensemble approach is simple and easy to implement, it is typically costly to work with many networks during training and inference. Here, we introduce a reliable end-to-end framework for the classification of Diabetic Retinopathy fundus images that combines the concepts of TTAug \cite{Wang_2019} and UA-Ensemble \cite{sarawgi2021uncertainty}, allowing us to obtain a reliable, well-calibrated final prediction by making the final ensemble of predictions aware of the model's inherent uncertainty. The novelty of our approach lies in it's ability to produce well-calibrated reliable results without compromising on model's efficacy. 

\begin{figure*}[t]
  \centering
  \includegraphics[width=0.69\textwidth]{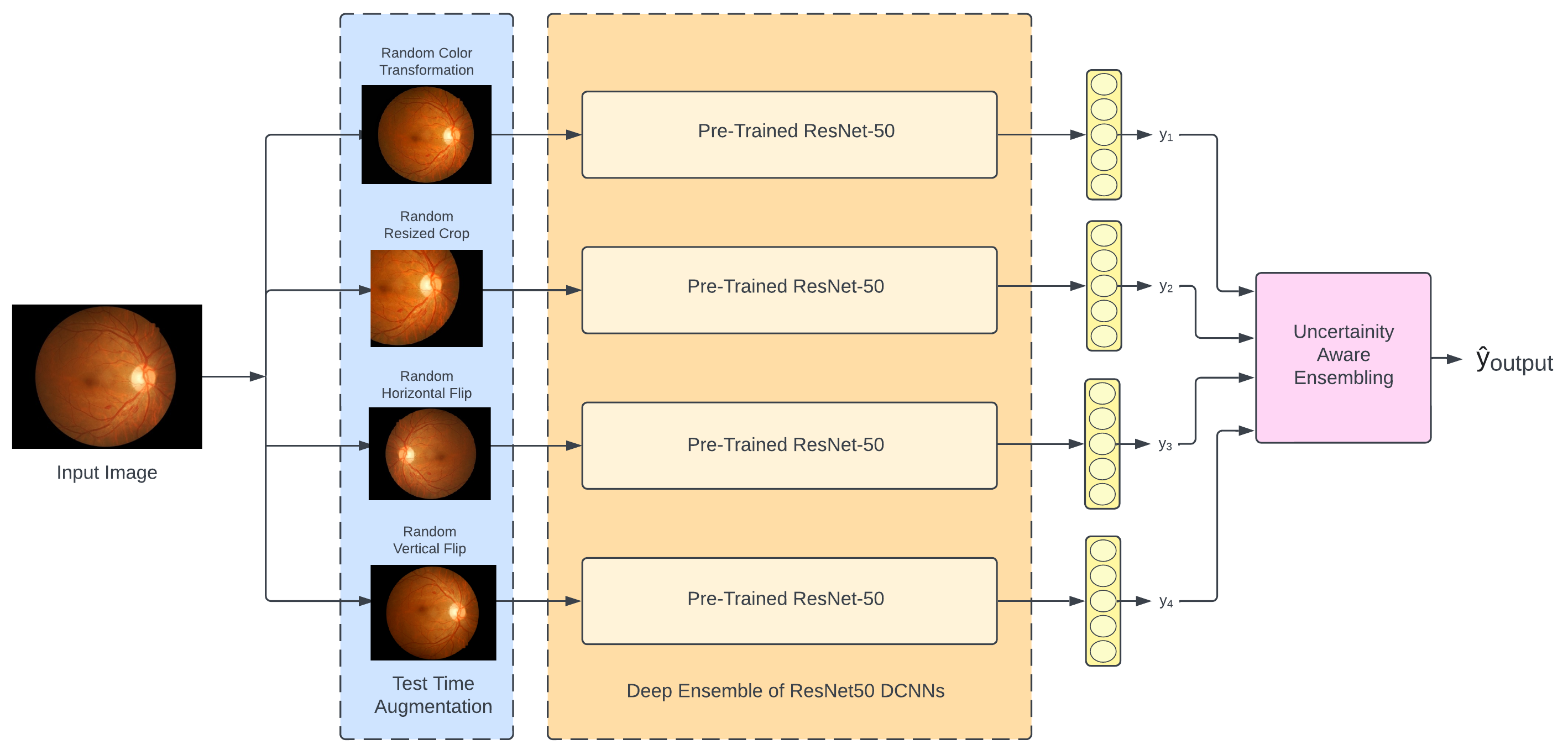}
   \caption{UATTA-ENS : Model Architecture}
   \label{fig:onecol}
\end{figure*}

\section{Methodology}
\subsection{Model Architecture}
\label{subsection:Model Architecture}

Given a training set
(x$_n$,y$_n$)$^N_{n=1}$
consisting of N i.i.d. samples, we model an end-to-end architecture as depicted in Figure \ref{fig:onecol}. 
The architecture is trained in two stages. In the first stage, four individual models are trained to classify the disease. In the second stage, we use TTAug \cite{Wang_2019} as described in section \ref{subsection:tta}. The pre-trained models receive the augmented images as input, which they then use to produce individual classification outputs. These individual classification outputs are then ensembled using calculated uncertainty as weights to produce the final output, or $\hat{y}$, and the entire architecture is fine-tuned over the training set.
\subsection{Test-Time Augmentation}
\label{subsection:tta} 
Test-time Augmentation is a technique used to improve the performance of image classification and produce well-calibrated uncertainty estimates \cite{Ayhan2020ExpertvalidatedEO}\cite{Wang_2019}. The train and test images are randomly augmented, which accounts for the noise, perturbation, and quality degradation that may arise in real-world data, thus making the model more robust. The transformations used in our approach include augmentation in brightness (-0.15 $<$ b $<$ 0.15), saturation (0.5 $<$ s $<$ 2.5), hue (-0.15 $<$ h $<$ 0.15), and contrast (0.5 $<$ c $<$ 1.5) values, cropping a random portion of the image and consequently resizing it to the original dimensions, random horizontal and vertical flipping, respectively. 
\subsection{Uncertainty Metrics}
\label{subsection:uncertainty}
We calculate the uncertainty estimation of the output of each model to curate an uncertainty-aware ensemble model. We use three metrics to quantify the uncertainty - Expected Calibration Error (ECE) \cite{nixon2019measuring}, Maximum Calibration Error (MCE) \cite{NEURIPS2019_1c336b80}, and Brier Score \cite{rufibach2010use}. More information about them is given in Appendix \ref{section:appA}.The output prediction $\hat{y_i}$ where $i\in(1, N)$ is weighted with its uncertainty. For the final prediction, $\hat{y}$, the weighted ensemble quantifies an uncertainty weighted average.
\begin{equation}
\hat{y}\left(\mathbf{x}_n\right)=\frac{\sum_{j=1}^k \frac{1}{\sigma_{h^j}\left(\mathbf{x}_n\right)} \hat{y}_{h^j}\left(\mathbf{x}_n\right)}{\sum_{j=1}^k \frac{1}{\sigma_{h^j}\left(\mathbf{x}_n\right)}}
\end{equation}
Here, $x_{n}$ denotes a nth input image, $\hat{y}_{h^j}(x_{n})$ is the output of nth input image for the $j_{th}$ model prediction, $\sigma_{h^j}$ is estimated uncertainty  corresponding to predictions from the $j_{th}$ model prediction. The uncertainty weights are formulated by taking an inverse of the independent uncertainty metric \cite{sarawgi2021uncertainty}. Hence, $\hat{y}\left(\mathbf{x}_n\right)$ outputs the final prediction corresponding to $n^{th}$ data point. Here the uncertainty associated with each prediction is quantified using a modified version of LLFU \cite{Lakara2021EvaluatingPU}.
\begin{equation}
\label{sec:llfu}
    \sigma_{h^j} = max(0,\frac{1}{2}log(2\pi\sigma^2(x_{n}))) 
    + 
    \frac{(y_j(x_{n})-\mu(x_{n}))^2}{2\sigma^2(x_{n})}
\end{equation}
where $y_j(x_{n})$ - denotes the prediction corresponding to $jth$ model, $\mu(x_{n})$ refers to the mode of predictions from all the models in ensemble and $\sigma^2(x_{n})$ refers to standard deviation of predictions of models in ensemble for the $n^{th}$ data point \cite{Jaskari2022UncertaintyAwareDL}. 

\begin{table*}[t]
\centering
\begin{tabular}{cccll}
\hline
Model Architecture       & Cohen-Kappa & ECE & MCE & Brier Score \\ \hline
Baseline(ResNet-50)        & 0.65            &0.25     &0.57     &0.27             \\
Ensemble         & 0.64            &0.15     &0.50     &0.16             \\
TTAug Ensemble     & 0.65             &0.17     &0.44     &0.17             \\
Uncertainity Aware Ensemble     & \textbf{0.68}            &0.16     &0.47     &0.18             \\
UATTA-ENS & 0.66            &\textbf{0.15}     &\textbf{0.29}     &\textbf{0.15}             \\ \hline
\end{tabular}
\caption{Test Accuracy and Uncertainty Metrics for various model architectures. }
  \label{tab:comb}
\end{table*}

\section{Experiments}
We conduct a series of experiments to study whether the architecture described in section \ref{subsection:Model Architecture} would be reliable in classifying the images. 
We use the APTOS 2019 Blindness Detection Dataset \cite{APTOS_DS}. The dataset contains labelled images of human retinas exhibiting varying degrees of Diabetic Retinopathy collected in India using different medical equipment. We use 90\% of the images (3,302 images) as a train set and the other 10\% (360 images) as a secondary validation set. The datasets consist of colored images of the human retina and are graded using the following 5-class PIRC system for the severity scheme of Diabetic Retinopathy. Each image was graded on the 0-to-4 scale \cite{Jaskari2022UncertaintyAwareDL}. \\ The images in the dataset have been resized to 512X512 and normalized. The black background of the images has been removed to focus more on the fundus image \cite{Huang2022IdentifyingTK}.
We have used four models in the ensemble. All the models used had ResNet-50 architecture with a sigmoid activation function used at the final linear layer. Each of these four pre-trained models was trained on random seeds and was ran for 50 epochs. We also use Stochastic Gradient Descent Optimization with a learning rate of 0.0001 and the weighted cross-entropy Loss to account for the severe class imbalance in the dataset.
\section{Results}
We evaluate the results over a fixed test set for the experiments. The DR grading performance is evaluated using Cohen’s quadratic weighted Kappa ($\kappa$) to measure the inter-rater agreement in ordinal multi-class problems \cite{warrens2013conditional}. More on this in Appendix \ref{section:appEVAL} 
We observe that all the models are quite similar in evaluation with Cohen Kappa Score as represented in Table \ref{tab:comb}. However, the uncertainty of these predictions is quite high. Also, most of these models are not well calibrated, leading to issues with the reliability models. Using our presented architecture by using TTAug \ref{subsection:tta} and Uncertainty Aware Ensemble \ref{subsection:Model Architecture}, we can reduce the uncertainty of our models by making the predictions much more reliable and well-calibrated, as shown in Table \ref{tab:comb}.
\section{Conclusion and Future Work}
Through the proposed model architecture, we aimed to make the model's predictions reliable and well-calibrated by making the model aware of uncertainty while ensembling predictions from multiple models and generalizing to distortions using TTAug \ref{subsection:tta}. We believe there is potential for this model architecture to be incorporated using Bayesian methods for various tasks involving classification and regression. We hope our work helps the community and inspires further research on uncertainty-aware learning of the model as well as making more reliable models for medical diagnosis.

\section*{Broader impact statement}
The approach introduced in this paper aims at tackling the issue of over-confident predictions during model outcomes for the diagnosis of Diabetic Retinopathy. Identifying Diabetic Retinopathy is one healthcare application where machine learning has already shown significant potential. However, its
applications in high-stakes healthcare choices must incorporate systematic uncertainty quantification and calibration for robust evaluation. By calibrating more generalized models, this framework also significantly aids in risk appraisal, lowering the possibility of potential misdiagnosis. We hope our work advances the current onset procedure for detecting Diabetic Retinopathy while also bringing trustworthy interpretations to other medical
imaging procedures.
\section*{Acknowledgement}
The authors would like to thank Mars Rover Manipal, an interdisciplinary student project team of MAHE, for providing the necessary resources for our research. We are also grateful to our faculty advisor, Dr Ujjwal Verma, for providing the necessary guidance.
\bibliography{references.bib}
\appendix

\section{Appendix} \label{apd:first}
\label{section:appA}


\subsection{Experiments}
\subsubsection{Evaluation Metric}
\label{section:appEVAL}
The DR grading performance is evaluated using Cohen’s quadratic weighted Kappa ($\kappa$) to measure the inter-rater agreement between raters in ordinal multi-class problems \cite{warrens2013conditional}. This metric penalizes discrepancies between ratings, which depend quadratically on the distance between the prediction and the ground truth, as follows:
\[ QCK = 1 -  
 \left[  \frac{\sum_{i}^C\sum_{j}^C w_{i,j}O{i,j}} { \sum_{i}^C\sum_{j}^C w_{i,j}E{i,j} }  \right]
\]
where C is the number of classes, w is the weight matrix, O is the observed matrix, and E is the expected matrix.

\subsection{Uncertainty Metrics}
We use three metrics to quantify the uncertainty - Expected Calibration Error (ECE) \cite{nixon2019measuring}, Maximum Calibration Error (MCE) \cite{NEURIPS2019_1c336b80}, and Brier Score \cite{rufibach2010use}.

\subsubsection{Expected Calibration Error}
The Expected Calibration Error (ECE)\cite{nixon2019measuring} is a weighted average over the absolute confidence difference of the predictions of a model. It is defined as 
\begin{equation}
\label{sec:ece}
 ECE =  \sum_{m=1}^M \frac{|B_m|}{n} |acc(B_m)-conf(B_m)|
\end{equation}
where
\begin{equation}
\label{sec:ece1}
  acc(B_m) =  \frac{1}{|B_m|} \sum_{i \in B_m}  1(y_{i}=y_{t})
\end{equation}
\begin{equation}
\label{sec:ece2}
 conf(B_m) =  \frac{1}{|B_m|} \sum_{i \in B_m}  p_{i}
\end{equation}
where $conf(B_m)$ is just the average confidence/probability of predictions in that bin, and $acc(B_m)$ is the fraction of the correctly classified examples B$_m$.
\subsubsection{Maximum Calibration Error}
The Maximum Calibration Error (MCE) \cite{NEURIPS2019_1c336b80} focuses more on high-risk applications where the maximum confidence difference is more important than the average.
It is then defined as:
\begin{equation}
\label{sec:mce}
  MCE =  max_{m} |acc(B_m)-conf(B_m)|
\end{equation}
\subsubsection{Brier Score}

The Brier Score \cite{rufibach2010use} is a strictly proper score function or scoring rule that measures the accuracy of probabilistic predictions.
\begin{equation}
\label{sec:bs}
  Brier Score = \frac{1}{n} \sum_{t=1}^{n} (f_{t} - o_{t})^2 
\end{equation}
where $f_{t}$ is the probability that was forecast, o{t} is the actual outcome of the event at instance t (0 if it does not happen and 1 if it does happen), and N is the number of forecasting instances.

\end{document}